\documentclass{article} 
\usepackage[final]{colm2026_conference}
\usepackage{microtype}
\usepackage{amsmath,amssymb}
\usepackage{hyperref}
\usepackage{url}
\usepackage{booktabs}
\usepackage{graphicx}
\usepackage{lipsum}
\usepackage{tikz}
\def\rvx{{\mathbf{x}}}

\def\rvz{{\mathbf{z}}}
\def\rvv{{\mathrm{v}}}
\def\rvq{{\mathrm{q}}}
\def\rvk{{\mathrm{k}}}
\newcommand{\mask}{\mathbf{m}}
\usetikzlibrary{arrows.meta,positioning,fit,backgrounds,calc}
\usepackage{xcolor,soul,xspace}
\usepackage{lineno}
\definecolor{darkblue}{rgb}{0, 0, 0.5}
\hypersetup{colorlinks=true, citecolor=darkblue, linkcolor=darkblue, urlcolor=darkblue}
\title{Retrofitting Linear Attention into Diffusion Language Models}
\author{%
Jinha Kim\thanks{Work done at MIT} \\ Apple \And
Younghun Roh\footnotemark[1]\\ Google DeepMind\And
Jaeyeon Kim\\ Harvard University
}

\begin{document}
\maketitle
\begin{abstract}
Diffusion language models (dLLMs) offer a promising alternative to autoregressive models by accelerating inference through parallel decoding. Recent dLLMs commonly use blockwise semi-autoregressive decoding, generating blocks autoregressively while denoising tokens within each active block in parallel. However, despite KV caching, each denoising step still attends to all previous blocks, repeatedly incurring prefix-attention cost. Motivated by this bottleneck, we ask whether dLLM inference can be further accelerated by linearizing attention over previous blocks. We introduce \emph{block-hybrid} attention, which retains exact softmax attention within the active denoising block while applying linear attention over previous blocks. We show that this hybrid attention can be retrofitted into a pretrained dLLM with minimal post-training: \textsc{LLaDA-Hybrid} replaces 6 of the 20 attention layers in LLaDA~2.1~\citep{bie2026llada2}, a 16B open-source dLLM, largely following LoLCAT~\citep{zhang2024lolcats}. The conversion takes only approximately $60$ hours while preserving benchmark performance: 72.0\% vs. 75.6\% on HumanEval, 63.0\% vs. 57.7\% on MBPP+, and 86.7\% vs. 88.3\% on CMATH. With a Triton implementation, \textsc{LLaDA-Hybrid} achieves up to $1.7\times$ higher decoding throughput and supports more concurrent requests before exhausting memory, showing that pretrained dLLMs can be efficiently linearized for faster inference. Our code is available at: \url{https://github.com/Diuven/LLaDA-Hybrid}.
\end{abstract}
\section{Introduction}
\label{sec:intro}
Diffusion language models (dLLMs) have recently emerged as a compelling alternative to autoregressive language models. Unlike autoregressive models, which generate sequences strictly from left to right, dLLMs begin inference from a fully masked sequence and iteratively reveal tokens at arbitrary positions. Their practical appeal is twofold: having flexibility in the inference order can support globally constrained generation and planning~\citep{trainin2026discrete,kim2025train,ye2024beyond}, and parallel decoding can substantially improve serving throughput, as demonstrated in recent large-scale deployments~\citep{nie2025llada,ye2025dream,gemini2025diffusion,labs2025mercury}.

Recent dLLMs commonly adopt blockwise semi-autoregressive decoding: blocks are generated sequentially, whereas tokens within the active block are denoised in parallel. This structure enables KV caching over completed blocks while retaining the throughput benefits of parallel decoding, offering a speed-up over autoregressive language model counterparts. 

In this work, we ask whether this efficiency advantage can be pushed further by \emph{retrofitting} dLLMs with linear attention~\citep{katharopoulos2020linear}. Linear attention has been extensively studied for large-scale autoregressive models since it makes the per-token attention cost no longer grow with the context length by replacing the softmax attention with a linearized kernel feature map and summarizing the prefix in a fixed-size recurrent state.

Applying linear attention to a dLLM, however, is not straightforward: bidirectional dependencies alter the hidden states, which in turn change the keys and values, preventing them from being cached. To this end, we propose \emph{block-hybrid attention}: retaining exact softmax attention within the active denoising block and applying linear attention over previously completed blocks. This preserves bidirectional interactions where they are needed while compressing the generated blocks into a fixed-size state. To our knowledge, this is the \emph{first} approach to apply linear attention distillation to a diffusion language model.

Empirically, we show that LLaDA~2.1-mini~\citep{bie2026llada2}, a pretrained 16B-scale dLLM, can be \emph{retrofitted} with hybrid linear attention using only a lightweight post-training budget. Following the recipe of LoLCATs~\citep{zhang2024lolcats}, our procedure consists of two stages. Stage 1 freezes the backbone and trains only the kernel feature maps and gates to match the attention output of each replaced layer. Stage 2 attaches LoRA~\citep{hu2022lora} and fine-tunes the hybrid model using the standard masked diffusion objective to revive downstream accuracy. The resulting model achieves up to a 1.7× end-to-end speedup over a strong baseline already optimized with the SGLang framework \citep{zheng2024sglang}, while largely preserving downstream performance.
\begin{figure}[t] \label{fig;main}
\centering
\includegraphics[width=\linewidth]{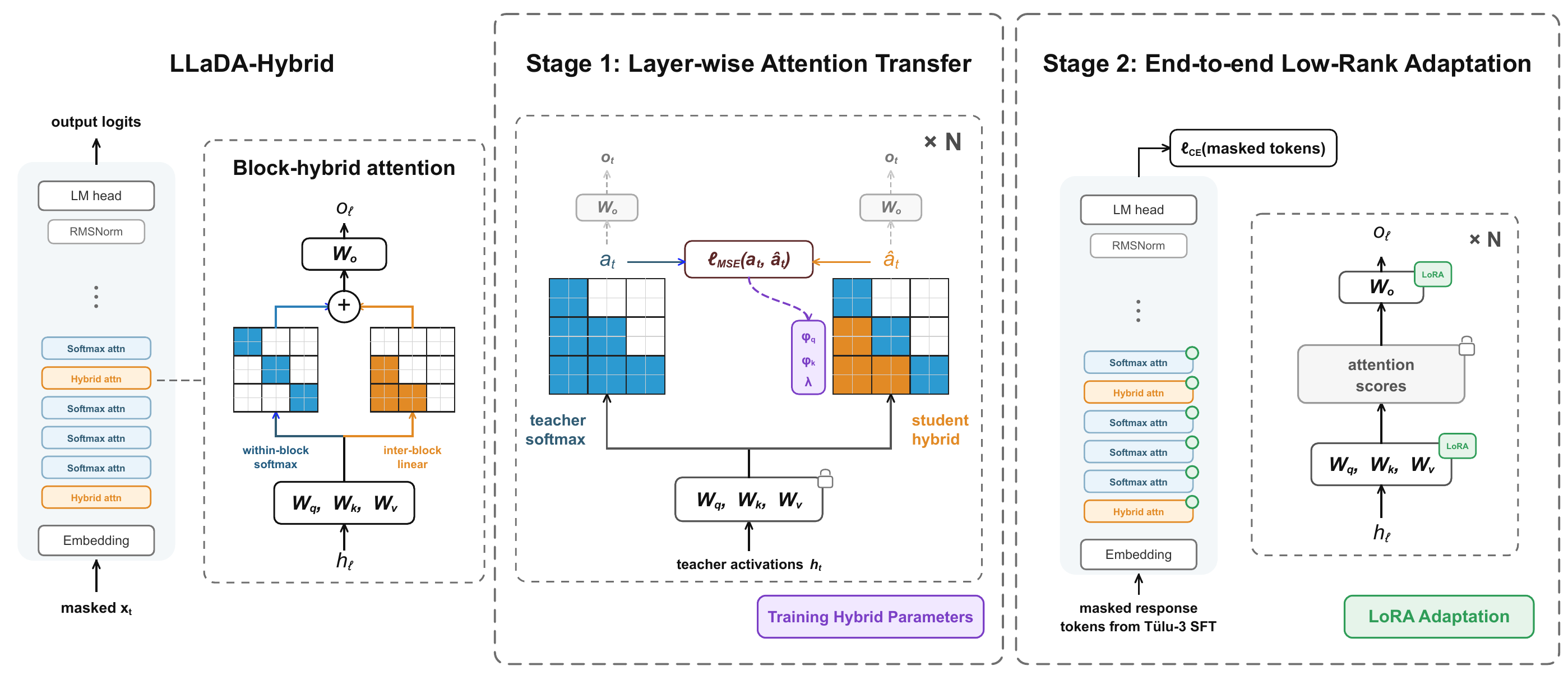} 
\caption{\textbf{Two-stage post-training.}
We first replace a subset of attention layers with block-hybrid attention.
In Stage~1, each hybrid layer is trained locally to match the corresponding frozen softmax-attention layer using the same corrupted activation input and an MSE attention-transfer loss.
In Stage~2, the full hybrid model is adapted on the masked-token diffusion objective with LoRA adapters, while the pretrained backbone remains frozen.}
\end{figure}
\section{Preliminaries}
In this section, we review diffusion language models and linear attention.

\subsection{Diffusion language models} Although there are other modeling approaches in discrete diffusion~\citep{sahoo2025diffusion,von2025scaling,schiff2025simple}, masked diffusion~\citep{sahoo2024simple,gat2024discrete,shi2024simplified} is currently the mainstream approach at scale~\citep{gemini2025diffusion,bie2026llada2,llada2scaling2025,lladamoe2025,ye2025dream,song2025seed}. LLaDA~2.1~\citep{bie2026llada2}, which serves as the main large-scale testbed of our paper, is also based on masked diffusion models (MDMs).

\textbf{Notation.} Suppose our goal is to learn the data distribution $\rvx\sim p_{\mathrm{data}}$ over length-$L$ discrete sequences with a vocabulary $\mathcal{V}$. Let $\rvx^i$ denote the $i$-th element of a given sequence $\rvx=(\rvx^1, \dots, \rvx^L)$ and $\Delta(\mathcal{V})$ indicate the probability simplex over $\mathcal{V}$.

\textbf{Training.} MDMs introduce an auxiliary mask token $\mask$ and learn, for each masked position, the posterior marginal of the clean token conditioned on a masked sequence. To learn this posterior, at training time, one draws a clean sequence $\rvx\sim p_{\mathrm{data}}$ and constructs a partially masked sequence $\rvz$ as follows: Sample $n\sim\mathrm{Unif}\{0,\dots,L\}$ and replace the tokens at uniformly selected $n$ indices in $\rvx$ with $\mask$. Hence, a resulting $\rvz$ has $n$ (randomly drawn) masked indices.
\begin{equation*}
    \mathcal{L}(\theta)\colon =  \mathbb{E}_{\rvx,\rvz}\left[\frac{1}{n}\sum_{i\colon \rvz^i=\mask} -\log f_\theta^i(\rvx^i \,|\,\rvz) \right].
\end{equation*}
This masking procedure yields a joint distribution over $(\rvx,\rvz)$, and we refer to its conditional marginal $\mathrm{law}(\rvx^i\mid \rvz)$ as the \emph{unmasking posterior}. This unmasking posterior is the central object in MDMs and is modeled by a neural network $f_\theta$, which gets $\rvz$ as input and outputs a $|\mathcal{V}|\times L$ tensor. Concretely, for each $i$, $f_\theta^i(\cdot\,|\,\rvz)\in\Delta(\mathcal{V})$, models the unmasking posterior $f_\theta^i(v\,|\,\rvz) \approx p(\rvx^i=v\,|\,\rvz)$. To train the network $f_\theta$, we minimize cross-entropy loss summed over all masked indices.

\textbf{Inference.} MDM inference starts from a length-$L$ masked sequence $\rvx_1 = (\mask,\dots,\mask)$ or generally a given prompt $\rvx_1 = ([\texttt{prompt}],\mask,\dots,\mask)$, and proceeds over a monotonically decreasing time grid $t_0=1>\dots>t_N = 0$. At each step $t_\ell$, given a partially masked sequence $\rvx_{t_\ell}\in(\mathcal{V}\cup \{\mask\})^L$, we proceed by two steps to attain $\rvx_{t_{\ell+1}}$: \textbf{(a)} Choose a subset of masked positions $\mathcal{S}\subseteq\{i\,|\,\rvx_{t_\ell}^i=\mask\}$ and \textbf{(b)} For each $i \in \mathcal{S}$, unmask $\rvx_{t_\ell}^i$ to a clean token $v$ sampled from $v\sim f_\theta^i(\cdot\,|\, \rvx_{t_\ell})\in\Delta(\mathcal{V})$.
Notably, as the MDM training is any-order, i.e., $f_\theta$ predicts the clean token distribution over all masked positions, there is flexibility in the choice of $\mathcal{S}$, which is central to the downstream performance.

Although parallel decoding can provide a speed-up, it still requires computing bidirectional attention over the full length-$L$ sequence. This can become a bottleneck, as bidirectional attention makes KV caching invalid. To address this, block diffusion~\citep{arriola2025block}, which is commonly adopted in diffusion language models, performs block-by-block autoregressive inference while applying parallel decoding within each block. This enables KV caching for previous blocks and therefore offers a speed-up.

\subsection{Linear attention} Softmax attention, the main design component of modern transformers, computes a normalized weighted average over all keys and values. For a single head with query $\rvq_i$, keys $\{\rvk_j\}_{j \le i}$, and values $\{\rvv_j\}_{j \le i}$, causal softmax attention takes the form
\begin{equation*}
    \mathrm{Attn}(\rvq_i) = \frac{\sum_{j\le i}\exp(\rvq_i^\top \rvk_j/\sqrt{d})\,\rvv_j}
         {\sum_{j\le i}\exp(\rvq_i^\top \rvk_j/\sqrt{d})}.
\end{equation*}
This operation is expressive and remains effective at scale, but its cost grows with the context length: at inference time, even with KV caching, every new query must be compared against all cached keys. Linear attention~\citep{katharopoulos2020linear} bypasses this by replacing the exponential kernel with a feature-map inner product; $\exp(q^\top k/\sqrt{d}) \to \phi(q)^\top \phi(k)$, where $\phi(\cdot)\in\mathbb{R}^{F}$ denotes a kernel feature map. Under this replacement, the linear attention's output can be rewritten as
\begin{equation}
\mathrm{LinAttn}(q_i)
=
\frac{\phi(q_i)^\top \left(\sum_{j\le i}\phi(k_j)v_j^\top\right)}
{\phi(q_i)^\top \left(\sum_{j\le i}\phi(k_j)\right)}.
\end{equation}
This yields an efficiency gain since the entire prefix can now be summarized by a fixed-size recurrent state:
\begin{equation*}
    S_i\colon=\sum_{j\le i}\phi(k_j)v_j^\top,\quad Z_i\colon=\sum_{j\le i}\phi(k_j).
\end{equation*}
After a token is processed, its key/value contribution can be summed into the recurrent state and reused for all future queries. Consequently, linear attention reduces the growing softmax KV cache to a constant-size state per layer and head, replacing sequence-length-dependent attention with a recurrent update.

\begin{figure}[t]
\centering
\includegraphics[width=0.95\linewidth]{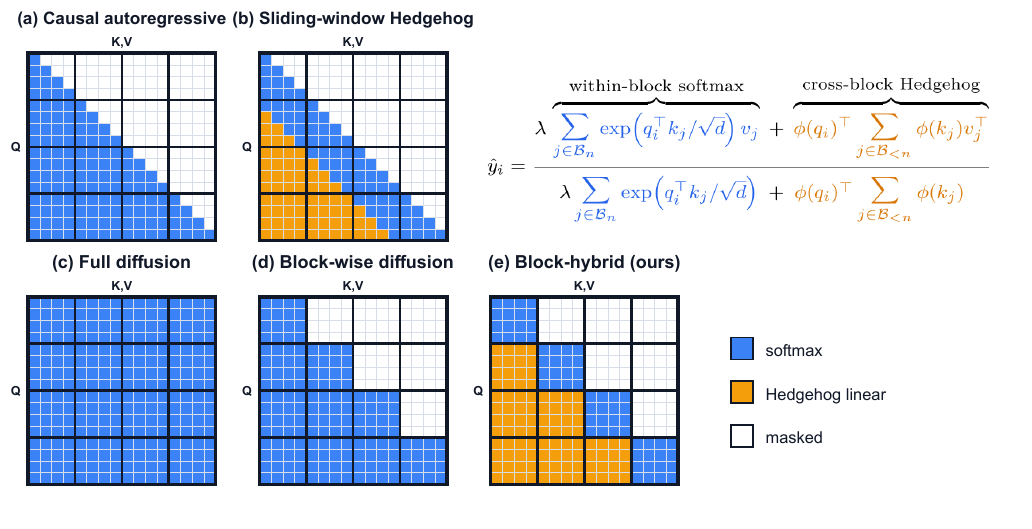}
\caption{\textbf{Attention-pattern comparison.}
Each panel shows a $16\times16$ query--key matrix split into four $4\times4$ blocks.
Blue is exact softmax, orange is Hedgehog linear attention, and white is masked.
\textsc{LLaDA-Hybrid} keeps the block-wise diffusion mask pattern, but uses softmax only within diagonal blocks and a fixed-state Hedgehog read for cross-block context.
The formula highlights the same split: within-block softmax in blue, cross-block Hedgehog in orange.
Orange cells are conceptual relationships represented by the linear branch, not a materialized dense matrix.}
\label{fig:attn}
\end{figure}
\section{Method: Block-hybrid attention} \label{sec:3.1}
While dLLMs are typically deployed using block diffusion with KV caching, standard softmax attention still incurs growing memory and computation costs as the sequence length grows. This motivates replacing part of the softmax attention computation with linear attention. Applying linear attention to dLLMs, however, is not straightforward due to their bidirectional nature; constructing a fixed-size recurrent state is therefore not direct.

To address this issue, we introduce \textbf{block-hybrid attention}, which separates attention into two components. Within the active block, we retain exact bidirectional softmax attention. Across blocks, we compress all previously committed keys and values into a fixed-size linear-attention state. This construction preserves the blockwise bidirectionality of dLLMs while making the cost of accessing the committed context independent of its length.

\subsection{Block-hybrid attention}
Consider generation over blocks $B_0,B_1,\dots$, where the tokens in the current block $B_{n+1}$ are denoised jointly and all preceding blocks $B_{\le n}$ have already been committed. For each query in the active block, block-hybrid attention combines two branches: (1) a bidirectional softmax-attention branch over tokens in the active block and (2) a recurrent linear-attention branch summarizing all previously committed blocks.

\emph{Within a block}, we compute exact softmax attention among the tokens in $B_{n+1}$. Let $q,k,v \in \mathbb{R}^d$ denote a query, key, and value for one attention head. For a query $q$ in the active block $B_{n+1}$, define the unnormalized attention
weight $a_j = \exp(q^\top k_j/\sqrt{d})$, the within-block numerator and normalizer are
\begin{equation*}
    \mathrm{sm\_num}(q)=\sum_{j\in\mathcal{B}_{n+1}} a_j v_j,\qquad\mathrm{sm\_den}(q)=\sum_{j\in\mathcal{B}_{n+1}} a_j.
\end{equation*}
Their ratio recovers exact softmax attention restricted to the active block.

\emph{Across blocks}, we summarize the keys and values from committed blocks using Hedgehog linear attention~\citep{zhang2024hedgehog}. In particular, for each query head, we use the learned positive feature map
\begin{equation*}
    \phi(x) = \big[\,\mathrm{softmax}(xW),\ \mathrm{softmax}(-xW)\,\big]\in\mathbb{R}^{2F},
\end{equation*}
where $W\in\mathbb{R}^{d\times F}$ is learned per query head. Following
Hedgehog, the positive feature map is designed to approximate the behavior of
the softmax attention kernel, while including both $xW$ and $-xW$ improves
its ability to represent the sign structure of dot-product similarities.

The committed context is represented by two recurrent states. After blocks $B_{\le n}$ have been committed, we define
\begin{equation*}
    S_n = \sum_{m \le n}\sum_{t \in \mathcal{B}_m} \phi(k_t)v_t^\top
\in\mathbb{R}^{2F\times d},\qquad Z_n = \sum_{m\leq n}
\sum_{t\in\mathcal{B}_m} \phi(k_t)\in\mathbb{R}^{2F}.
\end{equation*}
These states are updated once a block is committed:
\begin{equation*}
    S_n = S_{n-1} + \sum_{t \in \mathcal{B}_n} \phi(k_t)v_t^\top, \qquad Z_n = Z_{n-1}+ \sum_{t \in \mathcal{B}_n} \phi(k_t).
\end{equation*}

A query $q$ in the active block $\mathcal{B}_n$ accesses all preceding blocks through $\mathrm{lin\_num}(q) = \phi(q)^\top S_{n-1}$ and $\mathrm{lin\_den}(q) = \phi(q)^\top Z_{n-1}$. Crucially, the dimensions of $S_{n-1}$ and $Z_{n-1}$ do not depend on the number of committed tokens. Thus, the time and memory required for a query to access the cross-block context are $\mathcal{O}(Fd)$, independent of sequence length.

\textbf{Combining the two branches.}
We merge the exact within-block branch and the recurrent cross-block branch
using a shared normalizer. Let $w=\sigma(\alpha)$ be a learned scalar gate
for each query head. The final attention output is
\begin{equation*}
    \mathrm{out}(q)=\frac{w\,\mathrm{sm\_num}(q)+\mathrm{lin\_num}(q)}{w\,\mathrm{sm\_den}(q)+\mathrm{lin\_den}(q)}.
\end{equation*}
The shared denominator balances the attention mass assigned
to the active block and the committed context, rather than independently
normalizing the two branches and combining their already-normalized outputs.

Consequently, block-hybrid attention retains the bidirectional structure of
block diffusion while reducing the memory and computation associated with
cross-block attention. The only additional parameters in each linearized
layer are the per-head feature-map matrices $W$ and scalar gates $\alpha$.

\subsection{Retrofitting linear attention}
We now ask how to \emph{minimally} equip a pretrained model with linear attention. Developing an efficient approach is crucial, as pretraining a linear-attention model from scratch is typically expensive. We adapt the two-stage linearization recipe of LoLCATs~\citep{zhang2024lolcats} to the diffusion and block-hybrid setting. As illustrated in Figure~\ref{fig;main}, the first stage transfers the behavior of each softmax-attention layer to its block-hybrid replacement, while the second stage corrects the residual mismatch that emerges when the modified layers are composed end to end.

\textbf{Stage 1: attention transfer.} In the first stage, we freeze the pretrained backbone and optimize only the newly introduced block-hybrid parameters $\psi=\{W,\alpha\}$, where $W$ and $\alpha$ are defined in Section~\ref{sec:3.1}. The objective is to make each block-hybrid module reproduce the output of the softmax-attention module it replaces.

Let $a^{\mathrm{hyb}}_ \ell(\psi)$ denote the blended attention output of the block-hybrid module at layer $\ell$, measured before the output projection, and let $a^{\mathrm{sm}}_\ell$ denote the corresponding output of the frozen softmax teacher. We minimize
\begin{equation*}
    \mathcal{L}_{\mathrm{AT}}(\psi)
    \;=\; \sum_{\ell\in\mathcal{S}} \mathbb{E}_{x_t}
    \big\lVert\, a^{\text{hyb}}_\ell(\psi) - a^{\text{sm}}_\ell \,\big\rVert_2^2 ,
\end{equation*}
where $\mathcal{S}$ is the set of linearized layers and $x_t$ is sampled from the masked diffusion forward process. We discuss our design choice of $\mathcal{S}$ in Section~\ref{sec:exp}.

A key component of attention transfer is \emph{teacher forcing across layers}. Each block-hybrid layer receives the hidden states produced by the \emph{original softmax teacher} rather than the outputs of preceding linearized layers. Consequently, the loss at layer $\ell$ isolates the approximation error introduced by that layer. The only trainable parameters are $\psi=\{W,\alpha\}$.

\textbf{Stage 2: end-to-end adaptation.}
The layer-local attention transfer performed in Stage~1 does not guarantee that the complete hybrid model matches the teacher end to end. To correct the residual cross-layer mismatch, we perform a second stage of end-to-end adaptation using the original diffusion masked-token objective:
\begin{equation*}
    \mathcal{L}_{\mathrm{diff}} = \mathbb{E}_{x_0,t,x_t} \left[-\log p_\theta(x_0 \mid x_t)\right],
\end{equation*}
with the precise masking and loss conventions inherited from the pretrained dLLM. During this stage, we freeze both the pretrained backbone and the block-hybrid parameters $\psi=\{W,\alpha\}$, and instead attach low-rank adapters~\citep{hu2022lora} to the attention and MLP projections and optimize only the adapter parameters. Freezing $\psi$ preserves the attention behavior established in Stage~1.

The two stages therefore address complementary sources of error. Attention transfer minimizes the local approximation error of each replacement, whereas end-to-end adaptation corrects the distribution shift and error accumulation that arise when all block-hybrid layers operate jointly.

\subsection{Serving}
We implement the block-hybrid module as a \emph{single} fused Triton kernel within SGLang~\citep{zheng2024sglang}. A single kernel launch computes the within-block softmax branch, the linear readout from the recurrent state, the gated merge, and the causal state update, without writing intermediate results to HBM. The matrix multiplications are mapped to Hopper tensor cores.

The recurrent state is stored per query head alongside SGLang's existing paged KV cache, allowing it to reuse the same slot allocator and continuous-batching scheduler. The key serving property is that this state has a \emph{fixed} size, independent of the number of tokens it has absorbed. Consequently, the per-request memory usage and per-pass cost of a linearized layer remain constant with respect to sequence length, whereas the teacher's KV cache grows with every committed token. Under a fixed memory budget, this allows the server to admit substantially more concurrent requests before exhausting available memory.

\section{Experiments} \label{sec:exp}
We evaluate whether a pretrained diffusion language model can be partially linearized at low training cost while preserving its generation quality and improving serving efficiency.

\textbf{Model and linearization configuration.} We retrofit the pretrained 16B LLaDA~2.1-mini~\citep{bie2026llada2} following the two-stage procedure described in Section~\ref{sec:3.1}. Of its $20$ attention layers, we linearize a strided subset of six layers with zero-indexed layer indices $\mathcal{S}=\{0,4,8,12,16,18\}$. The remaining $14$ layers retain standard softmax attention.

This $6$-of-$20$ configuration represents a deliberate quality--efficiency trade-off. Linearizing more layers increases the fraction of attention computation whose recurrent state has fixed size, but also introduces additional approximation error. Through careful ablation, we find that our choice yields clear serving gains while largely preserving the quality of the pretrained teacher. We refer to the resulting model as \textsc{LLaDA-Hybrid}.

\textbf{Training data and optimization.} Both retrofitting stages use the public Tulu SFT mixture~\citep{lambert2024tulu3}, which contains a diverse collection of instruction-following examples. We truncate each example to at most $2{,}048$ tokens and use dynamic batching, such that the number of examples in a batch varies with sequence length. We apply the masked diffusion corruption only to assistant-response tokens; prompt tokens remain unmasked, matching the conditional-generation setting used at inference time.

During Stage~1, we set learning rate decayed from $10^{-2}$ to $5\times10^{-3}$. Attention transfer processes approximately $17$ million tokens and requires approximately $24$ hours on two NVIDIA L40S GPUs.

During Stage~2, we freeze both the pretrained backbone and the block-hybrid parameters and optimize rank-$16$ LoRA adapters~\citep{hu2022lora} attached to the attention and MLP projections. We use LoRA scaling $\alpha_{\mathrm{LoRA}}=8$, dropout $0.05$, and a learning rate decay from $5\times10^{-5}$ to $10^{-5}$. The selected checkpoint is reached after approximately $300$ optimization steps, corresponding to roughly $5$ million tokens and six hours of training on two L40S GPUs.

\textbf{Evaluation protocol.} For quality evaluation, we compare \textsc{LLaDA-Hybrid} with its LLaDA~2.1-mini teacher using the same no-edit decoding configuration with $\tau=0.7$. We evaluate on coding, mathematical reasoning, and scientific reasoning benchmarks.

For serving evaluation, we measure end-to-end decoding throughput using SGLang continuous batching on a single NVIDIA H200 GPU. We use a pool of $1{,}024$ prompts sampled from Alpaca-cleaned, fix the generation length to $2{,}048$ tokens, and sweep the number of concurrent requests. Additional isolated kernel and memory-scaling measurements are reported in the appendix.

\textbf{Quality Retention.} Table~\ref{tab:quality} compares \textsc{LLaDA-Hybrid} with the original softmax-attention teacher. Despite replacing six of the model's $20$ attention layers, the hybrid model largely retains the teacher's coding and mathematical-reasoning performance. The gaps on HumanEval, HumanEval+, and CMATH are between $1.6$ and $3.6$ percentage points, while performance is maintained on MBPP and improves by $5.3$ points on MBPP+. The largest degradation occurs on GPQA-Diamond, where the hybrid model trails the teacher by $8.1$ points.

These results indicate that a substantial subset of the model's attention layers can be retrofitted without broadly disrupting its pretrained capabilities, although the most difficult reasoning tasks remain more sensitive to approximation error.

\begin{table}[h]
\centering
\caption{Accuracy under no-edit decoding ($\tau{=}0.7$). Teacher: LLaDA~2.1-mini.
Student: \textsc{LLaDA-Hybrid} ($6$/$20$ layers linearized).}
\label{tab:quality}
\vspace{0.1in}
\begin{tabular}{lcc}
\toprule
Benchmark & Teacher (\%) & \textsc{LLaDA-Hybrid} (\%) \\
\midrule
HumanEval     & 75.6 & 72.0 \\
HumanEval+    & 72.0 & 68.9 \\
MBPP          & 70.4 & 70.6 \\
MBPP+         & 57.7 & 63.0 \\
CMATH         & 88.3 & 86.7 \\
GPQA-Diamond  & 38.9 & 30.8 \\
\bottomrule
\end{tabular}
\end{table}

\textbf{End-to-End throughput.} Table~\ref{tab:tps} reports decoding throughput under SGLang continuous batching. \textsc{LLaDA-Hybrid} is faster than the softmax-attention teacher at every tested concurrency level, achieving speedups between $1.50\times$ and $1.73\times$. The largest improvement occurs at $128$ concurrent requests, where throughput increases from $2{,}310.2$ to $3{,}994.4$ tokens per second.

The speedup becomes larger as concurrency increases from $16$ to $128$ requests. This behavior is consistent with the fixed-size recurrent state of the linearized layers: their cross-block attention does not require reading an increasingly large KV cache as generation progresses. At $256$ concurrent requests, the speedup remains substantial at $1.60\times$, although both models begin to encounter additional system-level bottlenecks.

\begin{table}[h]
\centering
\caption{Decode throughput (tokens/s, higher is better) under SGLang continuous batching on one
H200, generation length $2048$, $1024$-prompt Alpaca-cleaned pool.}
\label{tab:tps}
\vspace{0.1in}
\begin{tabular}{rccc}
\toprule
Concurrent requests & Teacher (tok/s) & \textsc{LLaDA-Hybrid} (tok/s) & Speedup \\
\midrule
16  & 1845.7 & 2761.7 & $1.50\times$ \\
32  & 2047.5 & 3101.9 & $1.51\times$ \\
64  & 2457.9 & 4026.4 & $1.64\times$ \\
128 & 2310.2 & 3994.4 & $1.73\times$ \\
256 & 2350.3 & 3767.5 & $1.60\times$ \\
\bottomrule
\end{tabular}
\end{table}

\textbf{Kernel efficiency and concurrency scaling.}
The same fixed-state property also reduces the memory required for long-context generation and increases the number of requests that can be served under a fixed memory budget. The appendix provides isolated microbenchmarks of the \textsc{LLaDA-Hybrid} attention kernel against SGLang's radix/paged-KV attention kernel, including kernel latency, memory usage, and the resulting concurrency limits (Table~\ref{tab:kern},Table~\ref{tab:maxbatch4096}).
\section{Conclusion}
We showed that pretrained diffusion language models can be retrofitted with linear attention without retraining from scratch. Our approach, block-hybrid attention, maintains 
the bidirectional softmax within each denoising block and deploys a fixed-size recurrent state that summarizes committed blocks. Linearizing 6 of 20 layers in LLaDA~2.1-mini preserves performance on coding, mathematics, and reasoning benchmarks while enabling up to $1.7\times$ higher decoding throughput and greater concurrency under a fixed memory budget. The conversion requires only public instruction data and on the order of $10^7$ training tokens, suggesting that linear-attention efficiency can be added to existing dLLMs as an inexpensive post-hoc retrofit.

\textbf{Limitations and future work.}
Our retrofit is intentionally conservative: only 6 layers are linearized, so the remaining layers retain sequence-dependent attention cost and KV memory. More aggressive or adaptive layer selection may yield a better quality--efficiency tradeoff. Our evaluation is currently restricted to moderate generation lengths comparable to those used during retrofitting. Evaluating long-form generation and long-context reasoning, together with longer-sequence attention transfer or methods specialized for this regime~\citep{zhang2024lolcats,liu2025lawcat}, is an important next step. The current model also assumes a fixed inference block size, motivating training procedures and architectures that remain robust across block sizes.

\bibliography{main}
\bibliographystyle{colm2026_conference}
\appendix
\section{Additional Benchmarking}
\label{app:bench}
\paragraph{Standalone attention-kernel microbenchmark.}
We isolate the per-block attention cost from end-to-end serving (scheduling, KV-cache
management, sampling) by microbenchmarking a single dLLM-block forward of the attention
kernel \emph{in isolation} on one H200. Each timed call is the latency of generating
\emph{one $32$-token block} for a request that already has $P$ tokens of context.

We compare the softmax attention (SGLang's radix/paged-KV kernel) against the \emph{exact production} \textsc{LLaDA-Hybrid} kernel
served by SGLang (per-query-head, bf16-WGMMA fused path), reported by batch size in
Tables~\ref{tab:kern} (median of $20$ timed iterations).

\begin{table}[h]
\centering
\caption{Per-block kernel latency on one H200. Teacher uses exact attention over prefix length $P$; Hybrid reads a fixed-size recurrent state.}
\label{tab:kern}
\small
\begin{tabular}{rccc ccc}
\toprule
& \multicolumn{3}{c}{Batch $=8$} & \multicolumn{3}{c}{Batch $=32$} \\
\cmidrule(lr){2-4}\cmidrule(lr){5-7}
Prefix $P$
& Teacher (ms) & Hybrid (ms) & Speedup
& Teacher (ms) & Hybrid (ms) & Speedup \\
\midrule
$0$     & 0.090 & 0.070 & $1.3\times$  & 0.094 & 0.101 & $0.9\times$  \\
$256$   & 0.097 & 0.069 & $1.4\times$  & 0.147 & 0.100 & $1.5\times$  \\
$1024$  & 0.136 & 0.068 & $2.0\times$  & 0.319 & 0.103 & $3.1\times$  \\
$4096$  & 0.342 & 0.069 & $5.0\times$  & 1.114 & 0.101 & $11.0\times$ \\
$8192$  & 0.601 & 0.070 & $8.6\times$  & 2.150 & 0.103 & $20.9\times$ \\
$16384$ & 1.108 & 0.069 & $16.0\times$ & 4.204 & 0.101 & $41.8\times$ \\
$32768$ & 2.123 & 0.069 & $30.7\times$ & 8.299 & 0.103 & $80.6\times$ \\
\bottomrule
\end{tabular}
\end{table}

The teacher kernel scales \emph{linearly} in $P$; the retrofit kernel is essentially
\emph{constant} in $P$ (it reads a fixed-size state), so the speedup grows with prefix length bound as in the measured regime; to $\sim$$31\times$ at batch $8$ and $\sim$$81\times$ at batch $32$ for a
$32$K-token prefix. The hybrid kernel's latency is also nearly flat in batch size
($\approx$$0.07$\,ms at $B{=}8$, $0.10$\,ms at $B{=}32$), whereas the
teacher's grows with both $P$ and $B$.

\textbf{Protocol.}
We measure the maximum number of equal-length requests that can be resident
concurrently on a single NVIDIA H200. Each request has a 4096-token footprint,
constructed as a 4064-token prompt plus one 32-token diffusion block. For a
given batch size $B$, all $B$ requests are admitted in a single wave and decoded
together. We increase $B$ until the run fails with CUDA
out-of-memory, and define $B^\star$ as the largest successful batch size.

To make the capacity test tractable, we front-load the sequence length into the
prompt. This reproduces the memory footprint of a 4096-token resident request
while requiring only one diffusion block to be decoded, rather than running all
intermediate blocks of a full long generation. Thus, this experiment measures
resident memory capacity, not end-to-end generation latency.

Because the maximum resident batch depends on the serving runtime's memory
partition between preallocated cache/state storage and temporary execution
buffers, we tune this partition separately for each model and report the best
attainable $B^\star$ for that model. In SGLang, this partition is controlled by
\texttt{mem\_fraction\_static}, which determines the fraction of GPU memory
reserved for model weights and the cache/state pool. Reporting the best value
per model avoids tying the comparison to an arbitrary memory split. LLaDA-Hybrid supports about $1.3 \times$ as many concurrent requests.

\begin{table}[t]
\centering
\caption{Maximum single-pass concurrent batch on one H200, at a 4096-token
per-request footprint. The SGLang static-memory fraction is swept separately for
each model and the best successful batch is reported.}
\label{tab:maxbatch4096}
\begin{tabular}{lcc}
\toprule
Model & Best \texttt{mem\_fraction\_static} & Max.\ batch $B^\star$ \\
\midrule
LLaDA\,2.1-mini (teacher) & $0.85$ & $544$ \\
LLaDA-Hybrid              & $0.82$ & $\mathbf{704}$ \\
\bottomrule
\end{tabular}
\end{table}

\begin{table}[h]
\centering
\caption{EvalPlus pass@1 on the full $164$-problem HumanEval(+) set, threshold $0.7$, no-edit
(freeze-linear LoRA, step $200$, no dropout): effect of the number of linearized layers.}
\label{tab:evalplus}
\begin{tabular}{lcc}
\toprule
Model & HumanEval & HumanEval+ \\
\midrule
Teacher (full softmax)                       & 75.6\% & 72.0\% \\
\textsc{LLaDA-Hybrid}, 6 linear layers (LoRA) & 72.6\% & 67.7\% \\
\textsc{LLaDA-Hybrid}, 8 linear layers (LoRA) & 68.3\% & 64.0\% \\
\bottomrule
\end{tabular}
\end{table}

\textbf{Architectural Ablation.}
Teacher is the full-softmax LLaDA 2.1-mini; \textsc{LLaDA-Hybrid} variants replace $6$ or $8$ of the $20$ attention layers with per-query-head Hedgehog linear attention (same T\"ulu-3 data, same dataloader, freeze-linear LoRA at step $200$, \emph{no dropout}).  Linearizing more layers steadily erodes accuracy---going from $6$ to $8$ linear layers costs ${\sim}4$ points of HumanEval pass@1---so the production model uses the smallest count that preserves quality. The production checkpoint additionally trains the LoRA with dropout $0.05$, which accounts for the small differences from the no-dropout scores reported here.


\end{document}